\documentclass[11pt]{article}
\usepackage[paper=a4paper,left=3cm,top=2cm,right=2cm,bottom=3cm]{geometry}
\usepackage{amssymb}
\usepackage{color, colortbl}
\definecolor{orange}{rgb}{1,0.5,0}
\usepackage{graphicx}
\usepackage{subfigure}
\usepackage{multirow}
\usepackage{setspace}
\usepackage{lineno}

\begin{document}
\title{An Improved Approach for Contrast Enhancement of Spinal Cord Images based on Multiscale Retinex Algorithm}
\author{Sreenivasa Setty$^{1}$, N. K Srinath$^{2}$ and M. C Hanumantharaju$^{3}$   \\
\emph{$^{1}$Dept. of ISE, Don Bosco Institute of Technology, Bangalore, India}\\
\emph{$^{2}$Dept. of CSE, R. V College of Engineering, Bangalore, India}\\
\emph{$^{3}$Dept. of ISE, Dayananda Sagar College of Engineering, Bangalore, India}\\
\small{\emph{Email : mchanumantharaju@gmail.com}}
}

\maketitle

\begin{abstract}
This paper presents a new approach for contrast enhancement of spinal cord medical images based on multirate scheme incorporated into multiscale retinex algorithm. The proposed work here uses HSV color space, since HSV color space separates color details from intensity. The enhancement of medical image is achieved by downsampling the original image into five versions, namely, tiny, small, medium, fine, and normal scale. This is due to the fact that the each versions of the image when independently enhanced and reconstructed results in enormous improvement in the visual quality. Further, the contrast stretching and MultiScale Retinex (MSR) techniques are exploited in order to enhance each of the scaled version of the image. Finally, the enhanced image is obtained by combining each of these scales in an efficient way to obtain the composite enhanced image. The efficiency of the proposed algorithm is validated by using a wavelet energy metric in the wavelet domain. Reconstructed image using proposed method highlights the details (edges and tissues), reduces image noise (Gaussian and Speckle) and improves the overall contrast. The proposed algorithm also enhances sharp edges of the tissue surrounding the spinal cord regions which is useful for diagnosis of spinal cord lesions. Elaborated experiments are conducted on several medical images and results presented show that the enhanced medical pictures are of good quality and is found to be better compared with other researcher methods. 


\end{abstract}

\section{Introduction}

Image enhancement \cite{book01} plays a fundamental role in digital image processing where human experts make an important decision based on the imaging information. The objective of an image enhancement algorithm is to reconstruct the recorded image similar to that of the true picture. Medical image enhancement \cite{book02} is one of the key research fields for the researchers due to widespread use of medical images in the diagnosis of various lesions. In recent years, a lot of advancements has progressed in medical imaging area such as Magnetic Resonance Imaging (MRI) technology, Computed Tomography (CT) scan, X-ray and Ultrasound etc. Although, high field open MRI system offer good clarity images, visual inspection and objective evaluation of the resulting images have poor signal to noise ratio and increased scanning time. 

The CT scan technology offers dynamic imaging and is low cost with equally as fast as MRI imaging. CT scan imaging is best suited for analysis of skull fractures, brain nerves and blood supply to brain. In general, CT scan is preferred for the pathology detection in the human head since this kind of imaging provides detailed information of the head image far better than MRI image. However, CT scan is associated with enormous amount of radiation hazard and is not suggested for the soft tissue pathology. The X-ray imaging systems are easy, cheaper and faster compared to other imaging systems. However, the X-ray imaging systems are not suited for detection of small nodes as well as detailed analysis of the structures of the human body. The emerging nuclear based medical imaging technology such as Positron Emission Tomography (PET) offers number of advantages over conventional imaging systems. For instance, this imaging system is more precise and is capable of providing information on functioning of structures with details. However this imaging system is more expensive and needs lot of preparation for patients. 

The images obtained based on the imaging systems mentioned earlier are good in clarity, more details and are amenable for pathology detection. However, there is enough room left for incorporating image enhancement techniques into the medical image processing in order to improve the contrast, luminance and visual quality of an image. In medical image enhancement, spinal cord enhancement is the most popular since the spinal cord image details highlighted offer more information to diagnose a variety of spinal cord lesions. In the United States, every year about 10,000 people suffer from primary and metastatic spinal cord tumors. Among those with more than 90\% of the spine tumors are metastatic. 

Spinal cord tumors have an effect on people of all ages. However, they are common in young and middle aged adults. Nowadays, spinal cord imaging has been improved significantly and has been used enormously in spinal cord pathology due to the advancement of MRI technology. The spinal cord is encapsulated in a compact, bony structure, any unusual growth can set pressure on sensitive tissues and impair function. Spinal cord tumors may cause weakness, paralysis, numbness etc., on both sides of the body since the spinal cord is a narrow structure. The proposed multirate technique adapted into multiscale retinex image enhancement algorithm assists in the diagnosis of abnormal growths of tissues found in or near the spinal cord.  

This paper is organized as follows: Section 2 gives a brief review of previous work. Section 3 describes the proposed multirate multiscale retinex based contrast enhancement method. Section 4 provides experimental results and comparative study. Finally, conclusion arrived is presented in Section 5.    
      
\section{Previous Work}

The most popular conventional techniques used for medical image enhancement are intensity correction \cite{journ01}, histogram equalization \cite{conf01}, contrast stretching \cite{conf02} etc. Of these, histogram equalization and adaptive histogram equalization techniques offer good results for unimodal pictures but fails to produce better results for bimodal scenes. The intensity and gamma correction based image enhancement schemes heavily depend on input images and fail to adapt itself depending on changes in the image details. Qingyuan et al. \cite{conf03} proposed an improved multi-scale retinex algorithm for medical image enhancement. In this scheme, Y-component of medical image is separated into edge and non-edge area subsequent to RGB to YIQ color space conversion. The MSR technique has been used for the non-edge area in order to accomplish the medical image enhancement. However, this method provides satisfactory results for immunohistochemistry images but this approach may not provide inevitable results for other medical images. 

Natarajan et al. \cite{journ02} proposed a fusion of MRI and CT brain images by enhancement of adaptive histogram equalization. This scheme uses MRI images as foreground and CT scan images as a background followed by adaptive histogram equalization prior to image fusion. This method is mainly used in application such as human brain analysis for pathology detection. However, the reconstructed brain images presented in this work are generally not satisfactory. Although the authors claim that this method may be used for the analysis of human brain in pathology detection, experimental results confirm that the approach is not amenable. Tamalika et al. \cite{conf04} proposed a new medical image enhancement technique based on intuitionistic fuzzy set. This method uses intuitionistic fuzzy set with two uncertainities as compared with standard fuzzy set of single uncertainity. Although the author claims that the contrast enhancement achieved for medical images is far better compared to other intuitionistic fuzzy methods, window operation used in this algorithm produces halo artifacts in the reconstructed images.
 
Sundaram et al. \cite{journ03} proposed a histogram modified local contrast enhancement for mammogram images. This scheme offers better contrast enhancement and preserves local information in mammogram images. In addition, this technique leads in the detection of micro-classifications presence in the mammogram image. The enhancement approach used in this scheme is well suited for mammogram images but fails to produce satisfactory results for other medical images. The present authors has coded the same algorithm in Matlab and found that this work is not suited for images such as brain, spinal cord and retinal etc. Indira et al. \cite{conf05} developed an algorithm for medical image enhancement in two steps. The first step corrects contrast of an image and in the second step wavelet fusion has been applied for medical image enhancement. The restored images using this scheme are generally not satisfactory since the two steps adapted in this work over enhances the image. Therefore, this approach may not provide acceptable medical image enhancement. 

Cosnardo \cite{journ04} work on spinal cord pathology provides tips for radiologists while diagnosing a variety of lesions. Ronald et al. \cite{journ05} has presented contrast enhancement of spinal cord for a patient with cervical spondylotic myelopathy. However, this work provides a detailed report on specific patient without concern to other cases. The fast color image enhancement based on fuzzy multiscale retinex has been proposed by Chao et al. \cite{conf06}. This method uses an adaptive Gaussian mask in order to reduce the computation time and offers better image enhancement scheme compared to conventional Multi-Scale Retinex with Color Restoration (MSRCR) \cite{journ06}. The authors have claimed that in order to reconstruct the image, pixel average is computed for up sampled lower scale and the medium scale image. However, this approach introduces black spots in the final reconstructed image. The authors have not justified how high quality image enhancement has been achieved in spite of combining the upsampled  lower scale with that of medium scale image. 

The limitations mentioned earlier are overcome in the proposed method in an efficient way. The input image is first downsampled into five versions namely, tiny, small, medium, fine and normal scale. This downsampled versions are contrast stretched to increase the range of pixels and subsequently enhanced by MSR algorithm. After performing the enhancement, lower scale is upsampled to the size of next upper scale and then combined with the next upper scale image. While combining these images, if the upsampled image has a zero pixel, then the upper scale pixel is retained otherwise, the pixel average is computed. The proposed method removes the black spots present in the Chao et al. \cite{conf06} technique in an efficient way. The new modified MSR algorithm developed here is much faster compared to other existing methods since the image is downsampled into five versions. The multirate technique incorporated into multiscale retinex image enhancement algorithm is capable of enhancing spinal cord images. This is due to the fact that the pixel with zero value in upscaled tiny version of the image is replaced by the corrosponding pixel in the small scale version. Although a good contrast enhancement have been achieved for mammogram and retinal images, visual inspection and objective evaluation of the reconstructed images reveals that the proposed approach is best suited for spinal cord images. Therefore, the proposed work is tested with spinal cord MRI images and has a potential to diagnose a variety of lesions in these medical images.      

\section{Proposed Multirate Multiscale Retinex based Contrast Enhancement}

This section presents the proposed multirate MSR based contrast enhancement of spinal cord images. The flow sequence of the proposed method is presented in Fig. 1. Initially, the spinal cord images in dicom format are converted into ".tif" format in order to ease the complexity of processing. The original or normal scale RGB spinal cord image (".tif" format) with size $256\times256$ pixels is converted into Hue-Saturation-Value (HSV) color space since HSV space separates color from intensity. The value channel of HSV is scaled into five versions namely, tiny scale ($16\times16$ pixels), small scale ($32\times32$ pixels), medium scale ($64\times64$ pixels), fine scale ($128\times128$ pixels) and normal scale ($256\times256$ pixels) in order to speed up the MSR image enhancement algorithm. The hue and saturation components are preserved to avoid image distortion. Each of these scaled image versions may have a random pixel range. Therefore, contrast stretching operation is accomplished for each of the scaled versions of the value channel in order to translate the pixels to the display range i.e, 0 to 255. The contrast stretching operation a simple image enhancement method, improves the contrast of an image by stretching the range of original intensity values to span into a desired range of values. The contrast stretching is performed by using Eqn. \ref{eq1}

\begin{equation}
\label{eq1}
V_1(x, y) = \frac{d_{max}}{V_{Lmax}-V_{Lmin}}\left[V_L(x, y)-V_{Lmin}\right]
\end{equation}

where $d_{max}$ is the maximum intensity, which is chosen as 255 for an image with 8-bit representation, $V_L(x,y)$ is the value component of HSV image, $V_{Lmin}$  is the minimum value of value component, $V_{Lmax}$  is the maximum value of value component, $V_1(x, y)$ is the contrast stretched image and, x and y represents spatial coordinates.

\begin{figure}
\centering
{
\includegraphics[scale=0.15]{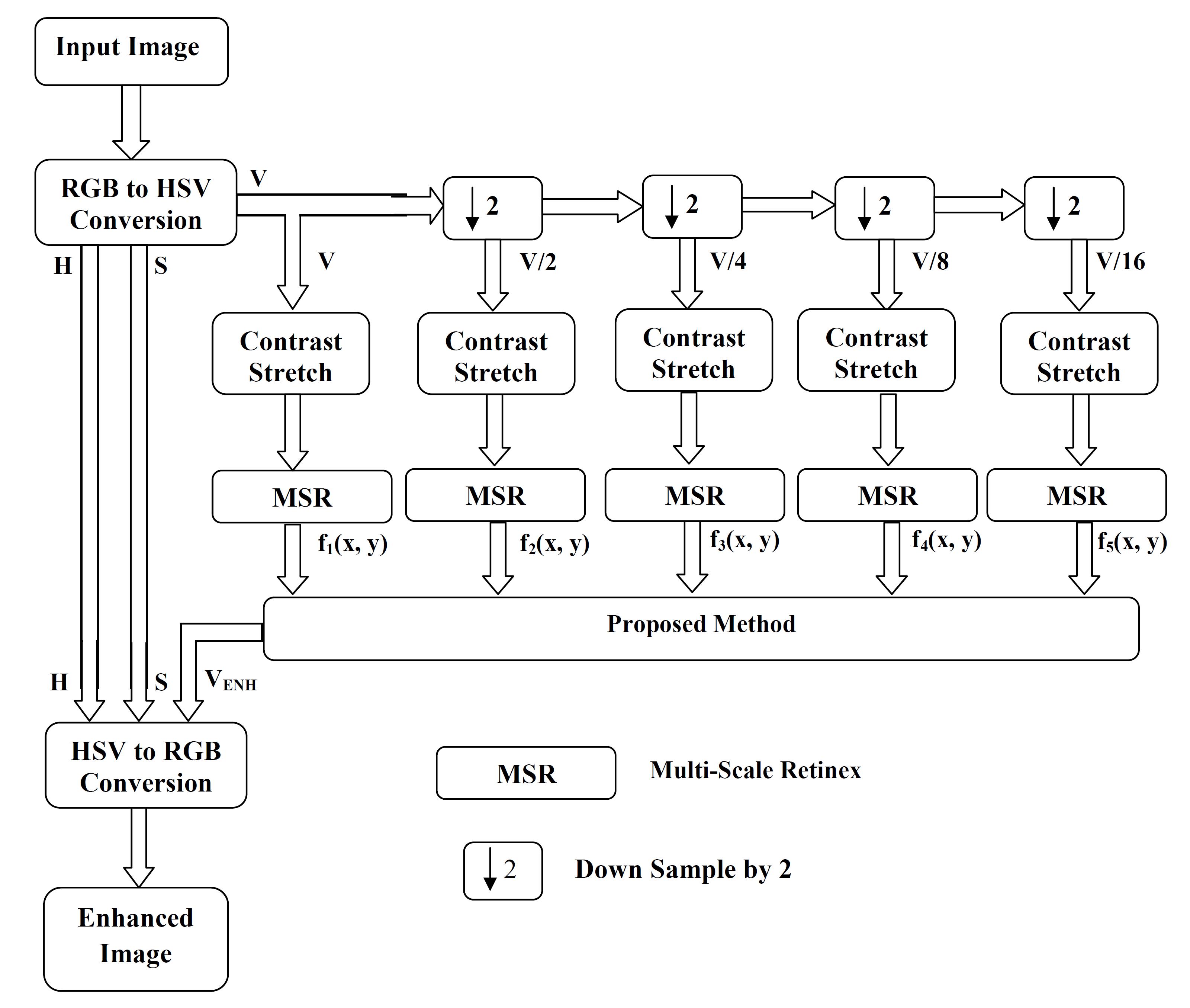}
\label{fig:subfig3}
}
\label{fig:subfigureExample}
\caption{Flow Sequence of the Proposed Multirate Multiscale Retinex based Contrast Enhancement}
\label{fig1}
\end{figure}

\subsection{The Multi-Scale Retinex Algorithm}

Numerous image enhancement algorithms presented earlier are used widely by many researchers. Among them, the most popular one is the MSR algorithm since this algorithm offers better image enhancement compared with other methods. Therefore, the present work adapts MSR technique in order achieve spinal cord medical image enhancement. The core of the MSR algorithm is the design of Gaussian surround function. The Gaussian surround function has been used in the proposed work for each scaled version of the contrast stretched value channel. The size of Gaussian function employed is $16\times16$ for the tiny version, $32\times32$ for the small version, $64\times64$ for the medium version, $128\times128$ for the fine version and $256\times256$ for the normal version of the value channel, respectively. The general expression for the Gaussian surround function is given by Eqn. \ref{eq2}.

\begin{equation}
\label{eq2}
G_n(x, y) = K_n \times e^{-\frac{(x^{2} + y^{2})}{2\sigma^{2}}}
\end{equation}

and $K_n$ is given by the Eqn. \ref{eq3}

\begin{equation}
\label{eq3}
K_n = \frac{1}{\sum_{i=1}^{M}\sum_{j=1}^{N}{e^{-\frac{(x^{2} + y^{2})}{2\sigma^{2}}}}}
\end{equation}

where x and y signify the spatial coordinates, $M \times N$ represents the image size, n is preferred as 1, 2 and 3 since the three Gaussian scales are used for each downsampled versions of the image.

Next, in order to accomplish medical image enhancement the SSR algorithm is follows MSR enhancement technique. The Single Scale Retinex (SSR) for the value channel is given by Eqn. \ref{eq4}

\begin{equation}
\label{eq4}
R_{SSRi}(x, y) = \log_2\left[V_i(x, y)\right] - \log_2\left[G_n(x, y)\otimes V_i(x, y)\right]
\end{equation}

where $R_{SSRi}(x, y)$ shows SSR output, $V_i(x, y)$ represents value channel of HSV, $G_n(x, y)$ indicates Gaussian Surround function and $\otimes$ denotes convolution operation. 

The Multi-Scale Retinex (MSR) operation on a 2-D image is carried out by using Eqn. \ref{eq5}

\begin{equation}
\label{eq5}
R_{MSRi}(x, y) = \sum_{n=1}^{N} W_n \times R_{SSRni}(x, y) 
\end{equation}

where $R_{MSRi}(x, y)$ shows MSR output, $W_n$ is a weighting factor which is assumed as $\frac{1}{3}$ and N indicates number of scales. 

The medical image enhancement is achieved by applying the MSR algorithm \cite{conf07} for each of the downsampled versions subsequent to SSR operation. The new value channel is reconstructed from the individual enhanced images by combining tiny, small, medium, fine and normal versions of the image in an efficient way. The MSR enhanced image of tiny version with resolution of $16\times16$ pixels is upsampled by two in order to match with the resolution of $32\times32$ pixels of the small version. However, the upsampling and reconstruction operations adapted by Chao et al. \cite{conf06} technique introduces zeros between alternative pixels. Although, an image enhanced by this scheme is satisfactory, it has actually resulted in appearance of dots in the enhanced image and thus affects overall image quality. The present work overcomes this difficulty in a proficient way. The new small scale version of the image is obtained as follows. The pixel of the small scale version is retained for the zeros encountered in the upsampled tiny version of the image. If there are no zeros in the upsampled tiny version image than the pixel average is computed between upsampled tiny version and small version. This is illustrated by the detailed flow chart presented in Fig. 2 (a) The new medium scale, fine and normal scale versions of the image is obtained in a similar manner as that for the small scale version and is shown in Figs. 2 (b), (c) and (d), respectively. Finally, the composite enhanced image is reconstructed by combining new value channel with that of hue and saturation channels and converting back into RGB color space.

\begin{figure*}
\centering
\subfigure[Flow Chart for Obtaining New Small Scale (Resolution of $32\times32$ pixels) of Enhanced Image]{
\includegraphics[scale=0.08]{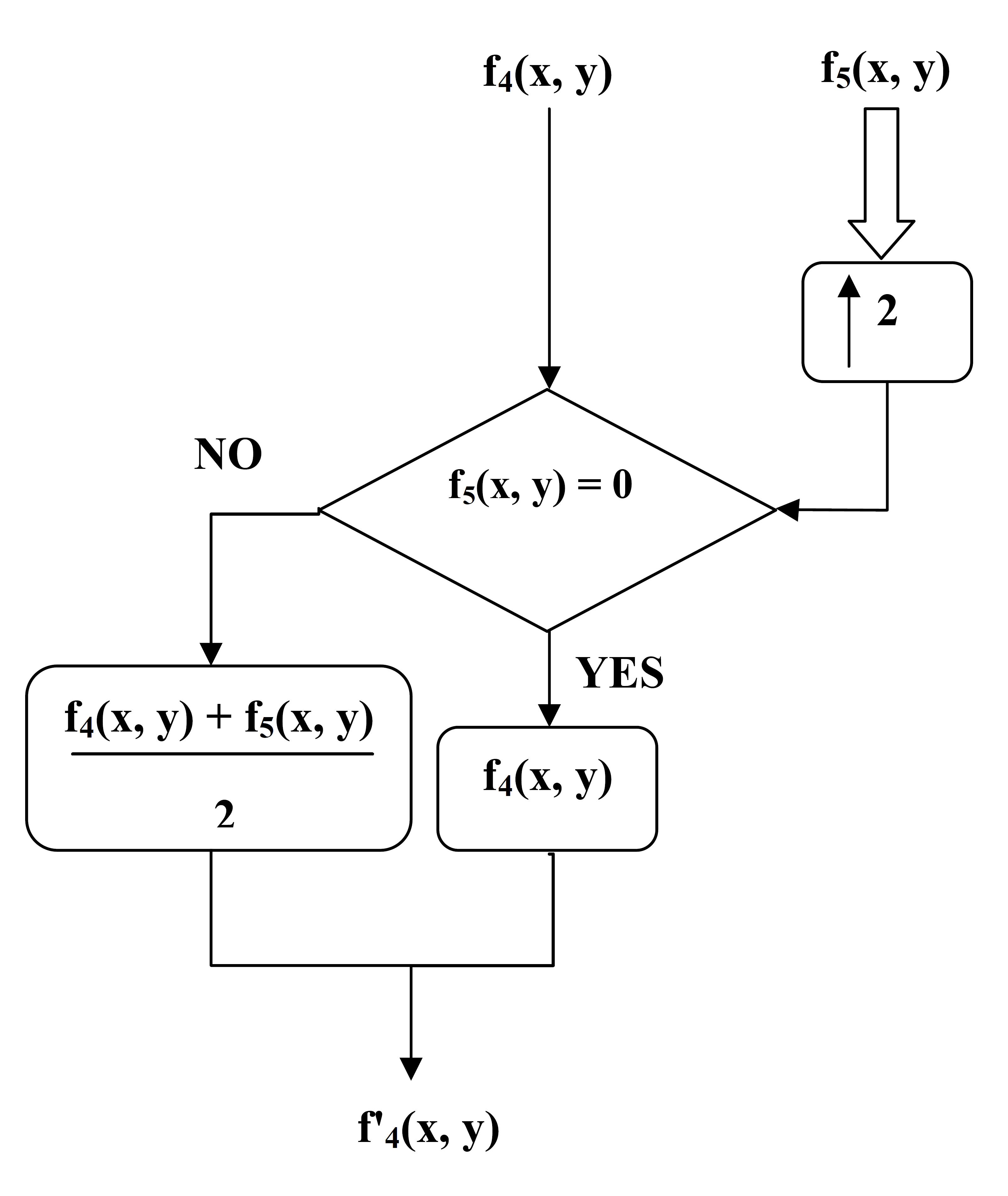}
}
\subfigure[Flow Chart for Obtaining New Medium Scale (Resolution of $64\times64$ pixels) of Enhanced image]{
\includegraphics[scale=0.08]{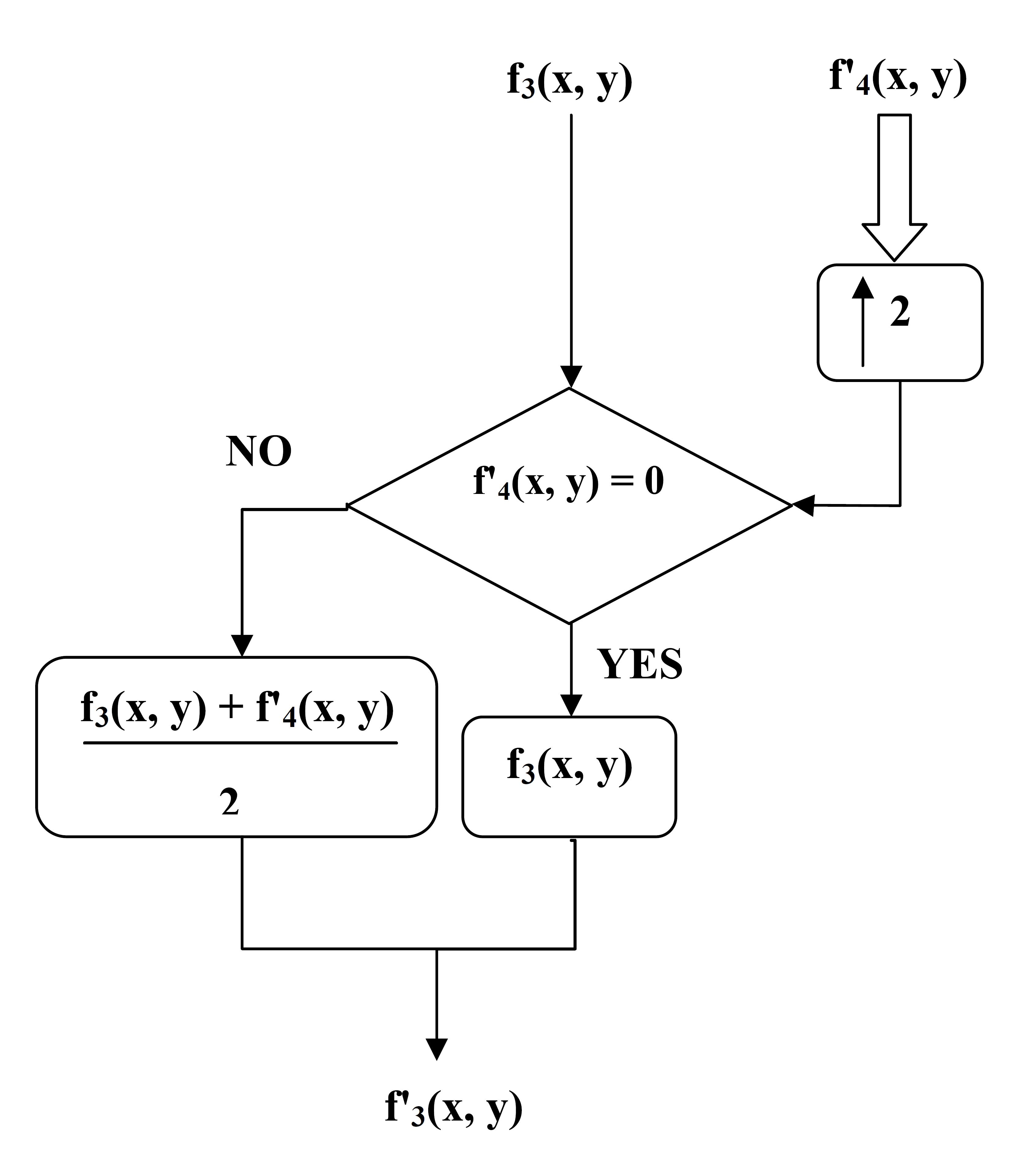}
}
\subfigure[Flow Chart for Obtaining New Fine Scale (Resolution of $128\times128$ pixels) of Enhanced image]{
\includegraphics[scale=0.08]{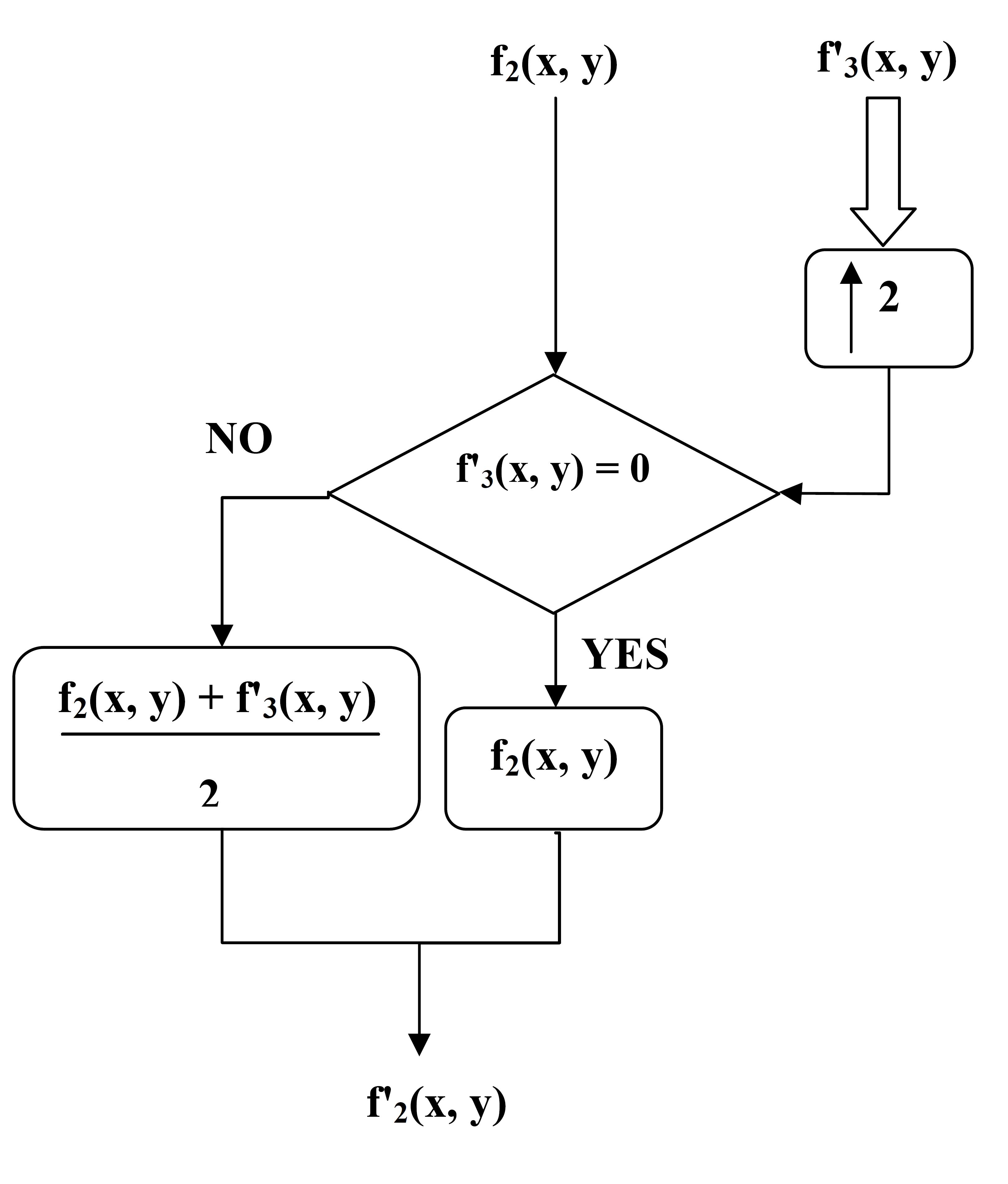}
}
\subfigure[Flow Chart for Obtaining New Normal Scale (Resolution of $256\times256$ pixels) of Enhanced image]{
\includegraphics[scale=0.08]{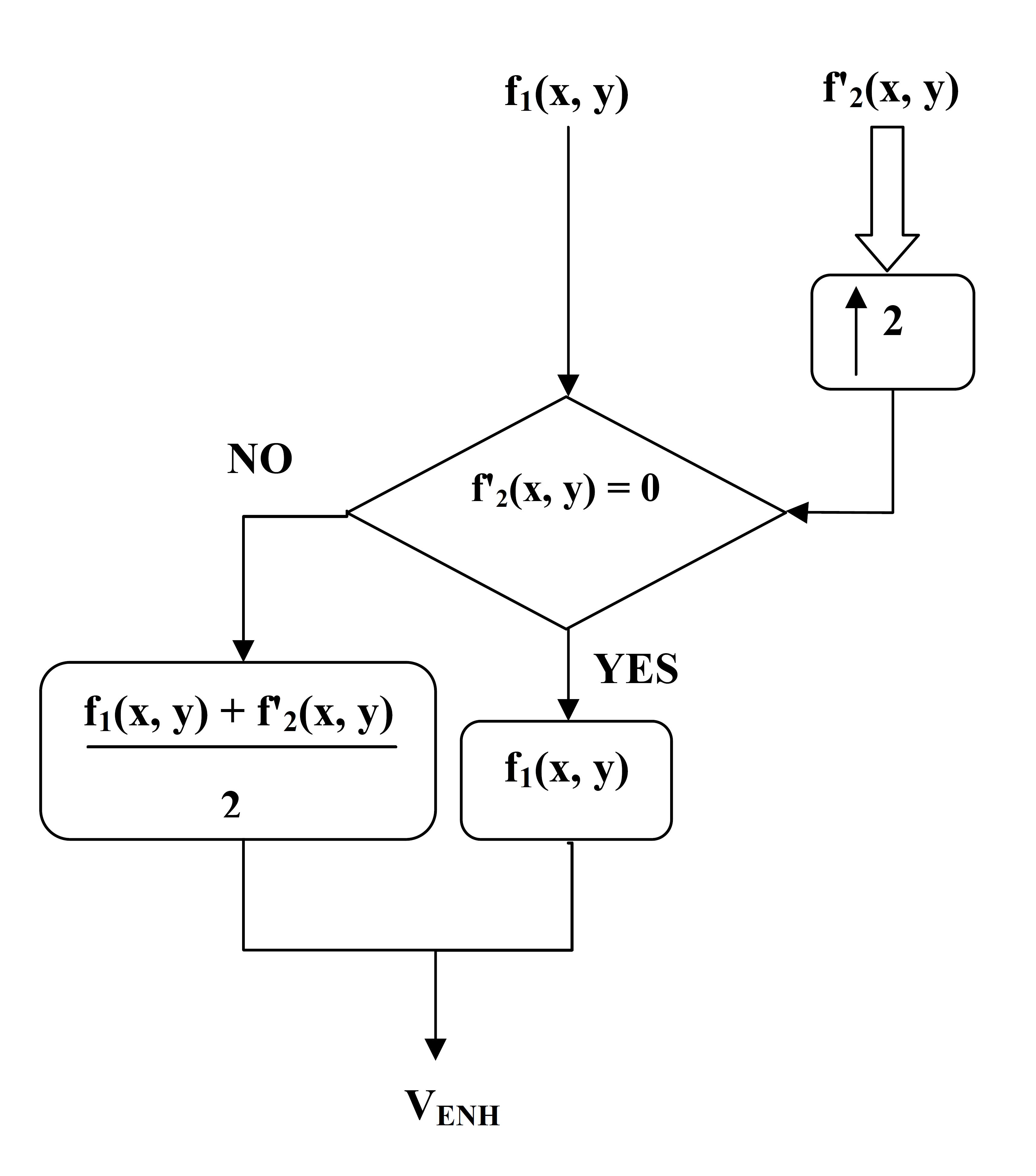}
}
\label{fig2}
\caption{Detailed Flow Sequence of the Proposed Multirate Multiscale based Spinal Cord Medical Image enhancement Technique}
\end{figure*}

The proposed algorithm uses HSV color space to enhance spinal cord images since this color space offers many advantages such as color separation from intensity, reduced color distortion and efficient enhancement. The algorithm proposed takes advantage of this color space by performing enhancment operation on the intensity or value component and preserving the color information. The first column of Fig. 3 shows the original spinal cord test images followed by the extraction of value component presented in second column. Next, the third column of Fig. 3 shows the value component enhanced using the proposed multirate multiscale retinex algorithm. The modified version of the retinex algorithm presented here is capable of producing high quality reconstructed pictures, far better than the other researchers method.

\begin{figure*}[ht]
\centering
\subfigure[Original Image]{
\includegraphics[height = 1.3 in, width = 1.5 in]{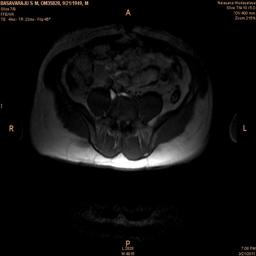}
}
\subfigure[Value Component of HSV]{
\includegraphics[height = 1.3 in, width = 1.5 in]{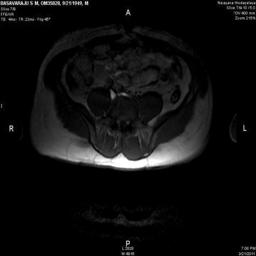}
}
\subfigure[Enhanced Value Component]{
\includegraphics[height = 1.3 in, width = 1.5 in]{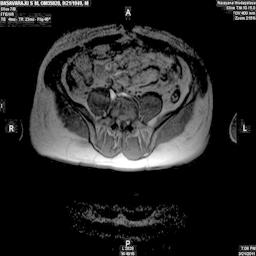}
}
\subfigure[Original Image]{
\includegraphics[height = 1.3 in, width = 1.5 in]{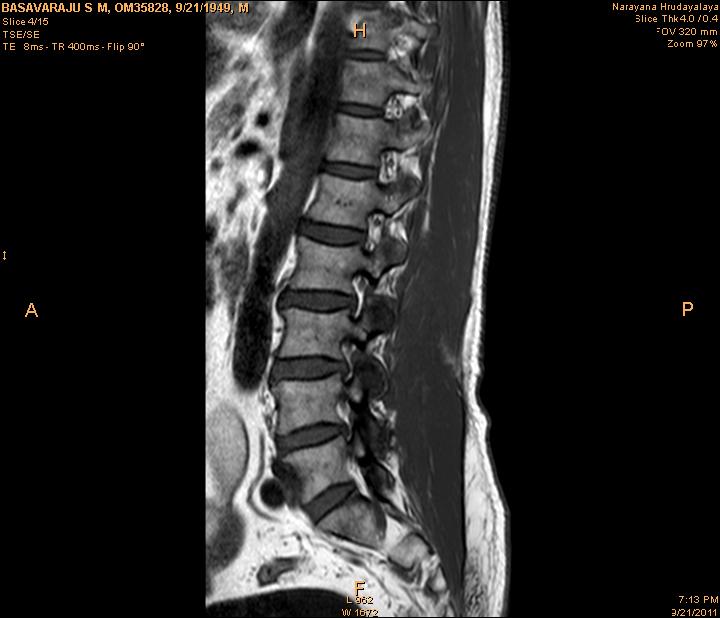}
}
\subfigure[Value Component of HSV]{
\includegraphics[height = 1.3 in, width = 1.5 in]{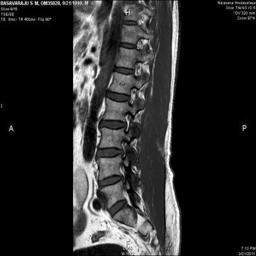}
}
\subfigure[Enhanced Value Component]{
\includegraphics[height = 1.3 in, width = 1.5 in]{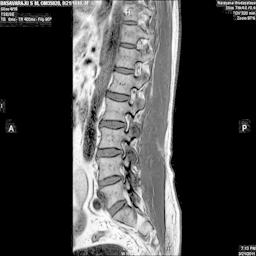}
}
\subfigure[Original Image]{
\includegraphics[height = 1.3 in, width = 1.5 in]{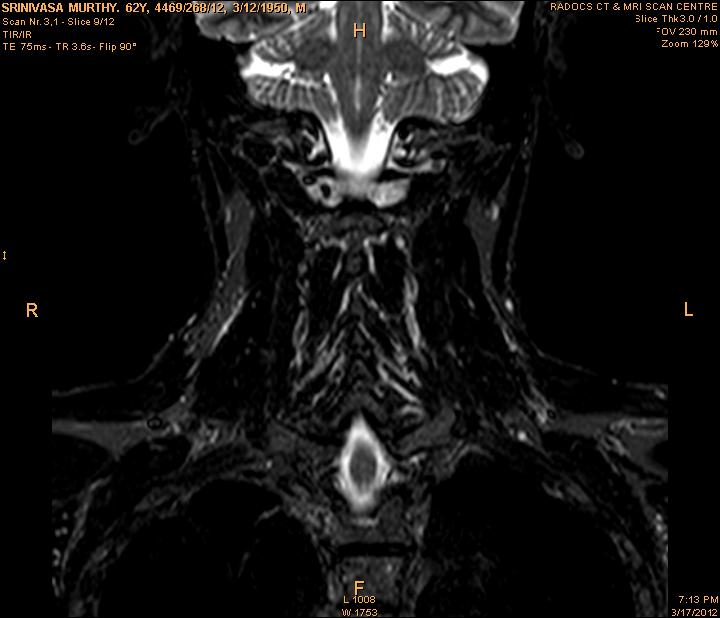}
}
\subfigure[Value Component of HSV]{
\includegraphics[height = 1.3 in, width = 1.5 in]{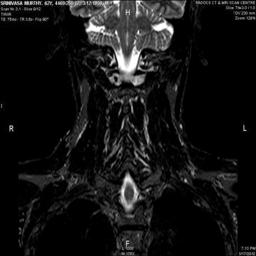}
}
\subfigure[Enhanced Value Component]{
\includegraphics[height = 1.3 in, width = 1.5 in]{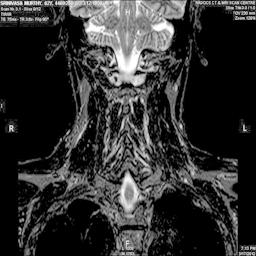}
}
\label{fig3}
\caption{Simulation Results of Reconstructed Value Component from the Value Component Extracted from the Original Image}
\end{figure*}

\section{Experimental Results and Comparative Study}

The developed algorithm presented in the previous section was coded using Matlab Version 8.0. The experiment was conducted by considering poor quality spinal cord images of having various lesions. The first column of Fig. 4 shows the axial view of neck spinal cord images of size $256\times256$ pixels, respectively. The second column of Fig. 4 presents the same images enhanced using histogram equalization. As is evident from the results presented, the histogram equalization method performs global image enhancement operation which improves the contrast of an image but at the cost loss in image details. The third column of Fig. 4 shows the MSR based spinal cord image enhancement. It can be seen that from the result presented, MSR scheme improves the details that are not clearly visible in histogram equalization technique. In general, image enhancement achieved by MSR method is better compared to histogram equalization. However, the MSR method voilates gray world assumption. Therefore, the image enhanced by this scheme appears to be grayish. Although, numerous work have been reported for solving the problem due to gray world voilation, no work seems to developed for complete elimination. 

The fourth column of Fig. 4 shows the image enhanced using Chao et al. []. It can be seen from the results presented that the reconstructed images of Chao's method have black spots. The appearance of these dark patches degrades the visual quality of the enhanced image. The image enhanced using proposed multirate multiscale retinex image enhancement method presented in the fifth column of Fig. 4, overcomes the drawback of the Chao's method. As we can see from the simulation results, image enhancement achieved by the proposed method has improved details with significant contrast enhancement. The enhanced images from the proposed method provides information to physicians, radiologists and researchers for various types of pathology detection. 

\subsection{Performance Evalution of Proposed Method using Wavelet Energy}
Wavelet Transform (WT) is a technique for analyzing the time frequency domain that is most suited for non stationary signal. The importance of WT is used to capture the localized features of the signal. A Continuous Wavelet Transform (CWT) maps a given function in time domain into two dimensional function of s and t. The parameter s is called the scale and corresponds to frequency in Fourier transform and t is the translation of the wavelet function. The CWT is defined by 

\begin{equation}
CWT(x,y) = \frac{1}{\sqrt{s}}\int S(T)\varphi\left(\frac{T-t}{s}\right)dt
\end{equation}

where S(T) is the signal and $\varphi(T)$ is the basic wavelet and $\varphi\left(\frac{T-t}{s}\right)\frac{1}{\sqrt{s}}$ is the wavelet basis function. The Discrete Wavelet Transform (DWT) is normally used for short time analysis. The DWT for a signal can be written as

\begin{equation}
DWT(m,n) = \frac{1}{2^{m}}\sum^{N}_{i=1}S(I,i)\phi\left[2^{-m}\left(i-n\right)\right]
\end{equation}

Wavelet energy is a method for finding wavelet energy for 1-D wavelet decomposition. The 1-D wavelet decomposition, wavelet energy provides percentage of energy corresponding to the approximation and the vector containing the percentage of energy corresponding details. The Wavelet Energy is computed as follows

\begin{equation}
WE = \frac{1}{2^{-m/2}}\sum^{N}_{i=1}S(I,i)\phi\left[2^{-m}\left(i-n\right)\right]
\end{equation}

In the proposed method, the WE is used as an effective and an efficient metric in order to evaluate the quality of the enhanced image. In most of the quality assesment metrics, high frequency information of an image is not sensitive to lightning changes or more prominent features, to descriminate different objects in an image. The WE is computed using a linear combination of high frequency coefficients after a Daubechies wavelet transform. The foreground object with poor contrast then the background of an image has continuous regular coefficient values, the Probability Density Function (PDF) for such image have relatively lower wavelet energy, lower skewness of the WE than a background. Thus the PDFs for such a poor contrast object is distinct when compared with the background. The wavelet energy computed for enhanced image and original image are compared and the following observations are made. Higher the detailed WE coefficients, details of the enhanced image is better. Alternatively, the details of the enhanced image are better for large detailed WE coefficients. However, the approximate WE coefficients provides information about the enhancement. Higher the approximate WE coefficients, globally the enhanced image is better. 

The wavelet energy based image quality assessment circumvents the discrepancies present in existing metrics and the human visual system. In the present work, the original and enhanced images are transformed into wavelet domain using DWT. It may be noted here that the wavelet transform, decomposes the image into four quadrants, namely, LL, HL, LH and HH. The upper left LL quadrant consists of low pass filtered wavelet coefficients at half the resolution of the approximated image. Next, the upper right and the lower left (HL/LH) block consist of vertical edges and horizontal edges of the image at different resolutions. The lower right (HH) block consists of high pass filtered coefficients. In this block, the edges of the image in the diagonal direction can be seen clearly. It may be analyzed that the standard deviation of the wavelet coefficients varies in accordance with the image quality. In accordance with this, the image quality in the proposed method can be measured by analyzing the characteristics of energy coefficients.

\begin{figure*}
\centering
\subfigure{
\includegraphics[height = 1.2 in, width = 1.1 in]{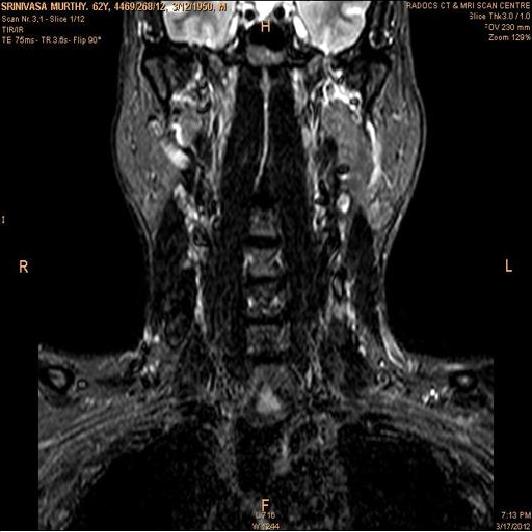}
}
\subfigure{
\includegraphics[height = 1.2 in, width = 1.1 in]{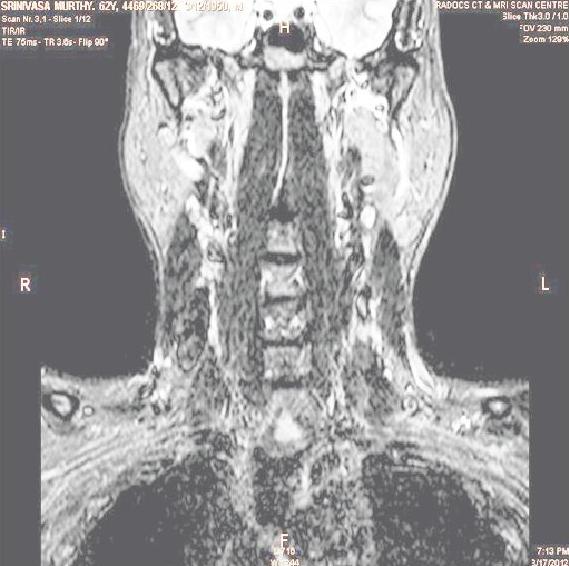}
}
\subfigure{
\includegraphics[height = 1.2 in, width = 1.1 in]{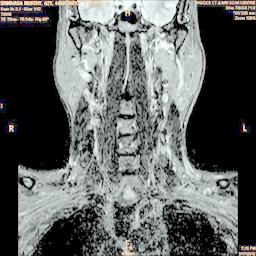}
}
\subfigure{
\includegraphics[height = 1.2 in, width = 1.1 in]{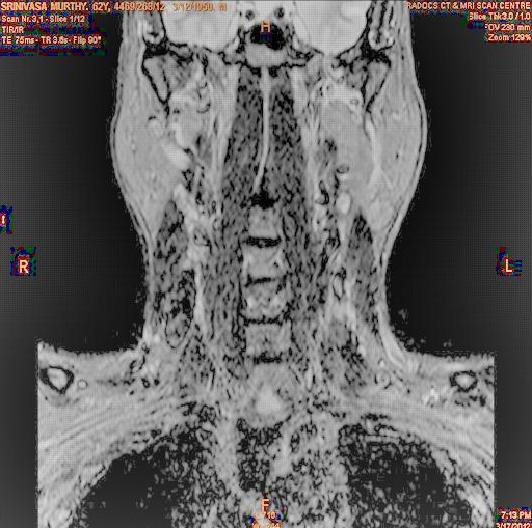}
}
\subfigure{
\includegraphics[height = 1.2 in, width = 1.1 in]{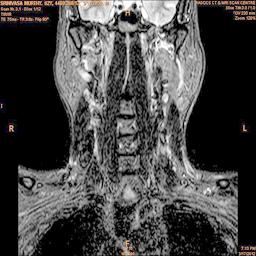}
}
\subfigure{
\includegraphics[height = 1.2 in, width = 1.1 in]{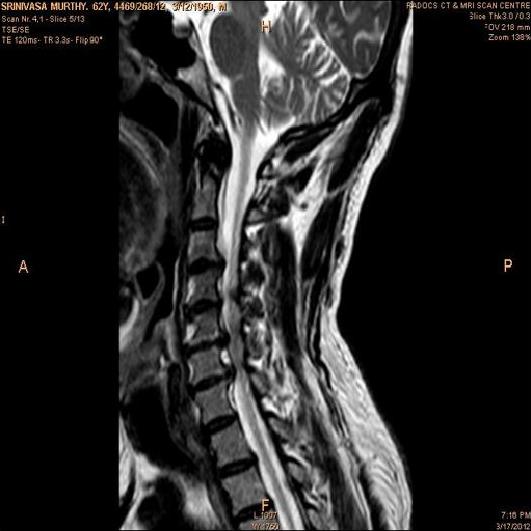}
}
\subfigure{
\includegraphics[height = 1.2 in, width = 1.1 in]{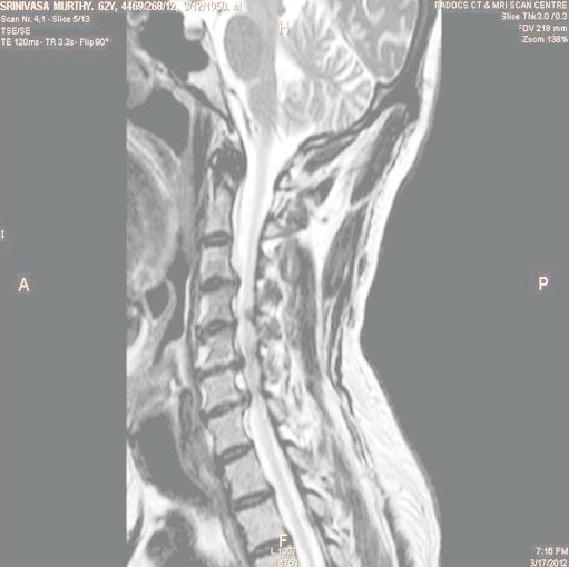}
}
\subfigure{
\includegraphics[height = 1.2 in, width = 1.1 in]{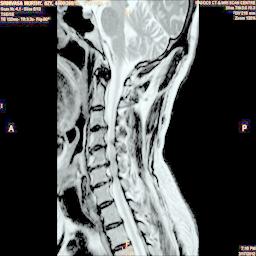}
}
\subfigure{
\includegraphics[height = 1.2 in, width = 1.1 in]{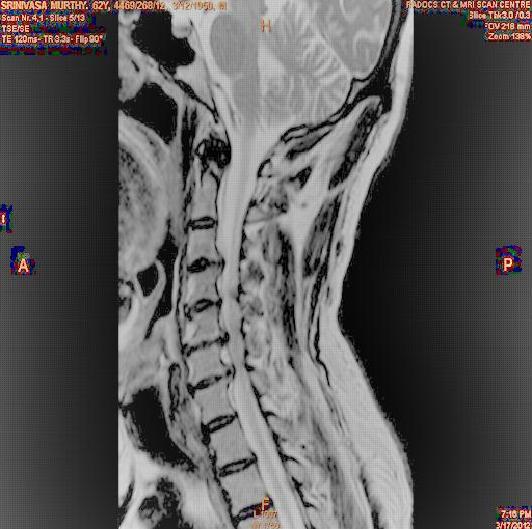}
}
\subfigure{
\includegraphics[height = 1.2 in, width = 1.1 in]{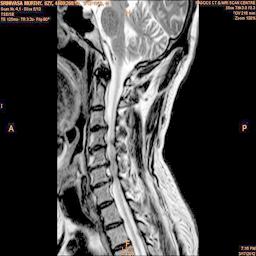}
}
\subfigure{
\includegraphics[height = 1.2 in, width = 1.1 in]{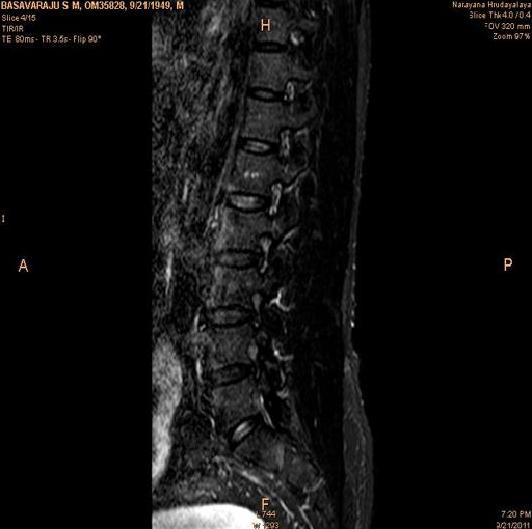}
}
\subfigure{
\includegraphics[height = 1.2 in, width = 1.1 in]{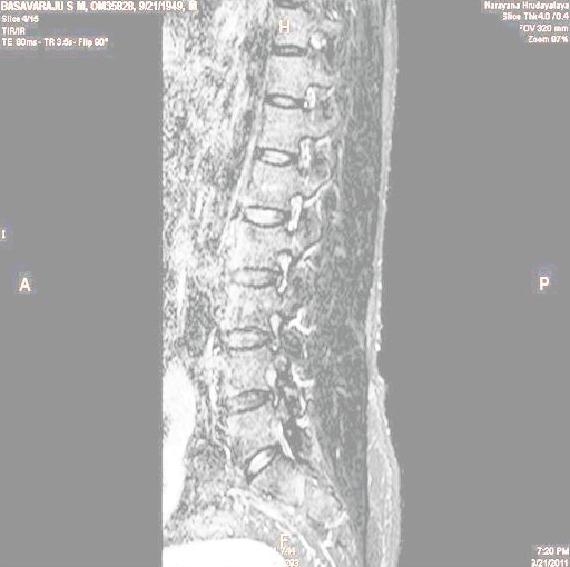}
}
\subfigure{
\includegraphics[height = 1.2 in, width = 1.1 in]{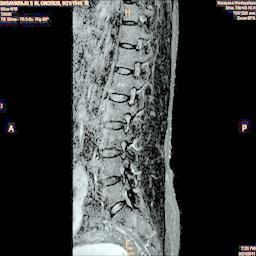}
}
\subfigure{
\includegraphics[height = 1.2 in, width = 1.1 in]{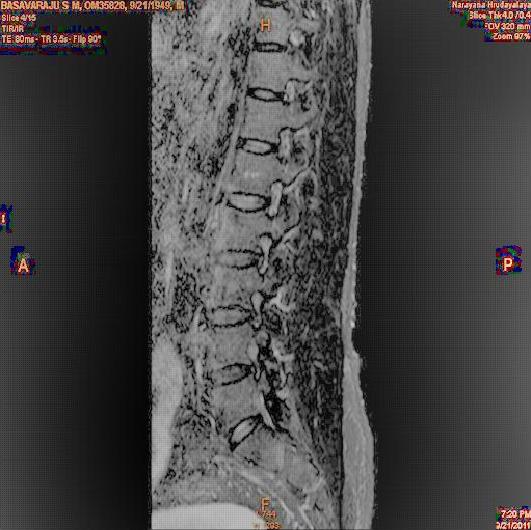}
}
\subfigure{
\includegraphics[height = 1.2 in, width = 1.1 in]{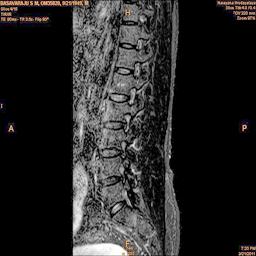}
}
\label{fig4}
\caption{Simulation Results Comparison with Chao et al. Work for a two Dimensional Axial View of Spinal Cord Images : \textbf{First Column:} Original Image of Resolution $256\times256$ pixels. \textbf{Second Column:} Image Enhanced using Histogram Equalization. \textbf{Third Column:} Image Enhanced using MSR Method. \textbf{Fourth Column:} Image Enhanced using Chao's work. \textbf{Fifth Column:} Image Enhanced using Proposed Multirate Multiscale Technique}
\end{figure*}

\begin{figure*}
\centering
\subfigure[Approximate Wavelet Energy]{
\includegraphics[height = 2.2 in, width = 3 in]{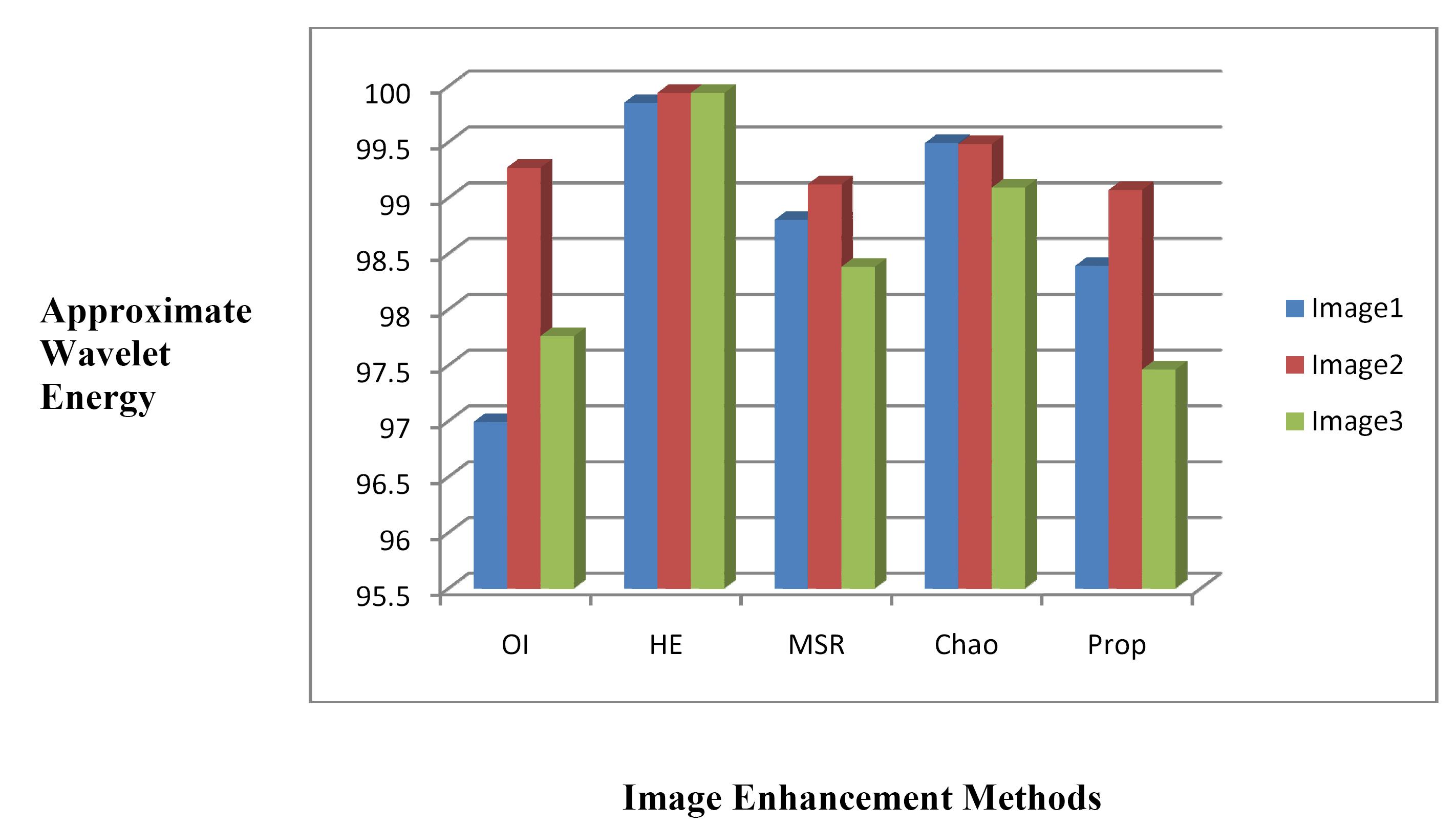}
}
\subfigure[Detailed Wavelet Energy]{
\includegraphics[height = 2.2 in, width = 3 in]{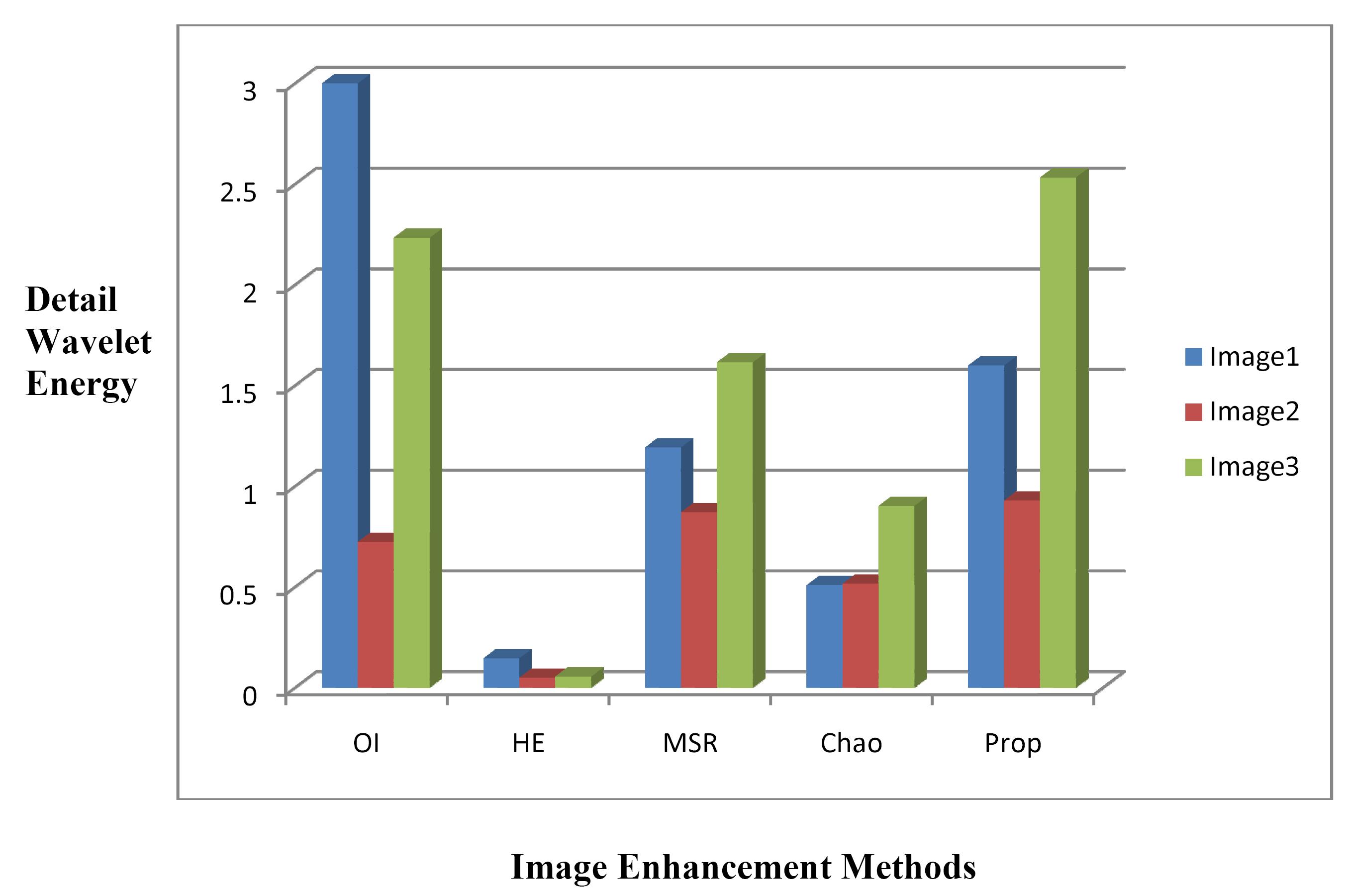}
}
\label{fig5}
\caption{Comparison of Image Enhancement Methods with Wavelet Energy based Image Quality Assessment for the reconstructed images shown in Fig. 4. \textbf{Note:} The wavelet energy computed for the reconstructed images are compared with the wavelet energy of the original image, followed by image quality assessment.}
\end{figure*}

The performance of the reconstructed spinal cord images are evaluated using the wavelet energy presented earlier is shown in Fig. 5. Fig. 5 (a) shows the Approximate Wavelet Energy (AWE) for the image enhancement methods such as histogram equalization, MSR, Chao's and proposed multirate multiscale image enhancement method. It can be seen from the AWE presented, higher the percentage of AWE of reconstructed image than the original image, better the image enhancement. Among image enhancement methods, histogram equalization technique has higher AWE coefficients than the original image owing to better global image enhancement. It may be mentioned here that the medical image enhancement inevitably needs more details rather than the global enhancement. The Detailed Wavelet Energy (DWE) coefficients shown in Fig. 5 (b) provides image details. Therefore, DWE of the reconstructed images are higher than the DWE of the original image. This shows that the proposed method offers better global enhancement and more detail enhancement. The image1, image2 and image3 of Fig. 5 corresponds to the images of first, second and third row of Fig. 4, respectively.

\begin{figure*}
\centering
\subfigure{
\includegraphics[height = 1.5 in, width = 1.4 in]{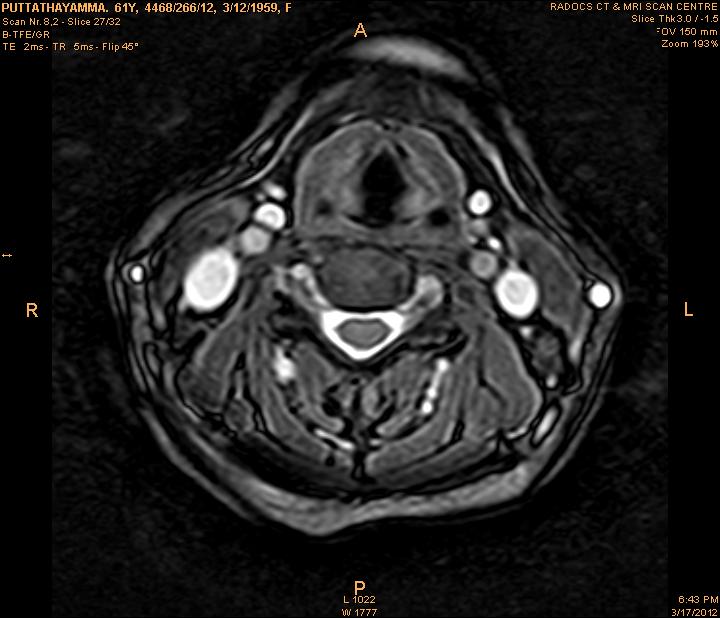}
}
\subfigure{
\includegraphics[height = 1.5 in, width = 1.4 in]{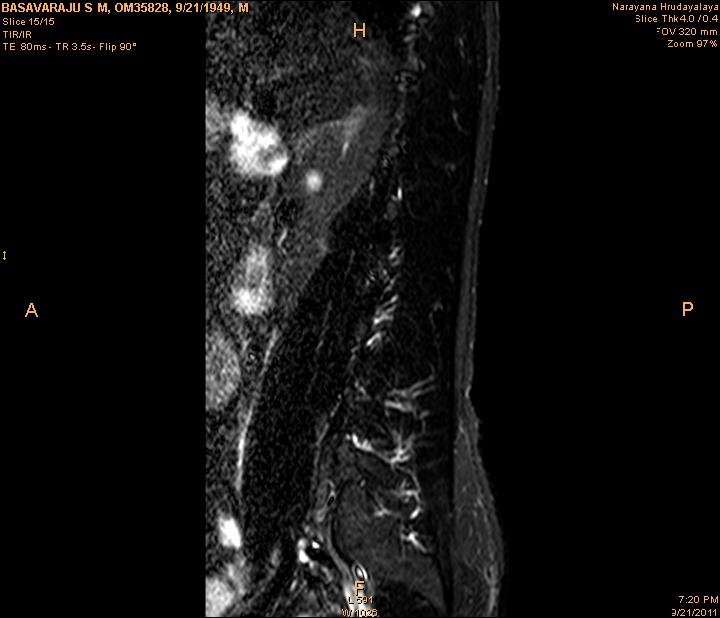}
}
\subfigure{
\includegraphics[height = 1.5 in, width = 1.4 in]{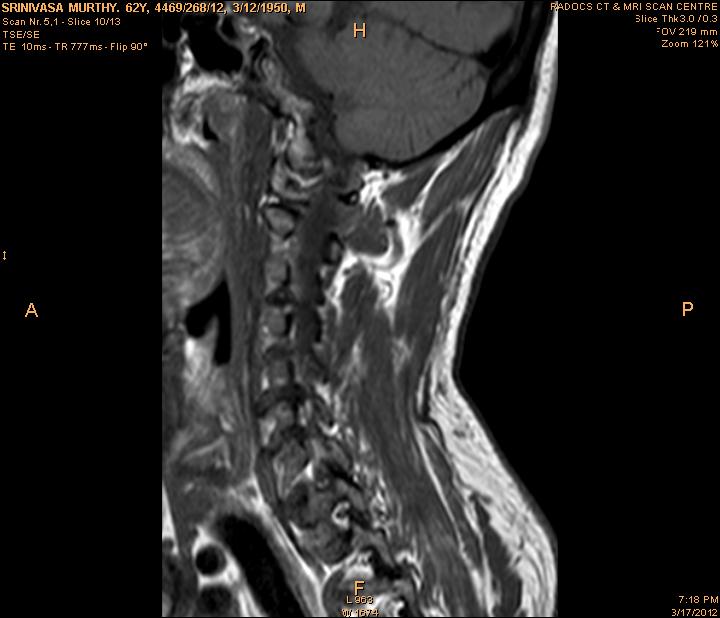}
}
\subfigure{
\includegraphics[height = 1.5 in, width = 1.4 in]{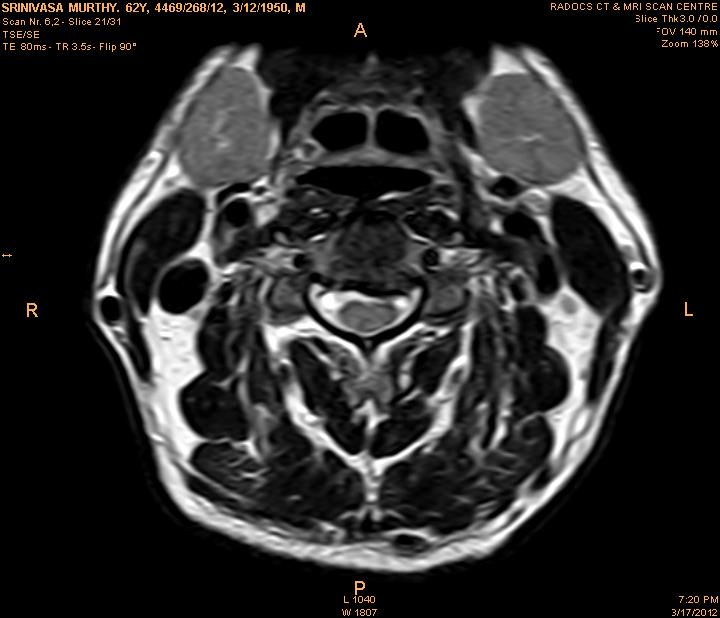}
}
\subfigure{
\includegraphics[height = 1.5 in, width = 1.4 in]{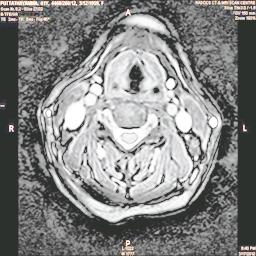}
}
\subfigure{
\includegraphics[height = 1.5 in, width = 1.4 in]{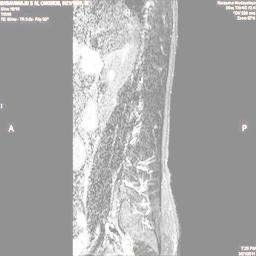}
}
\subfigure{
\includegraphics[height = 1.5 in, width = 1.4 in]{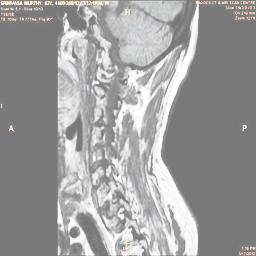}
}
\subfigure{
\includegraphics[height = 1.5 in, width = 1.4 in]{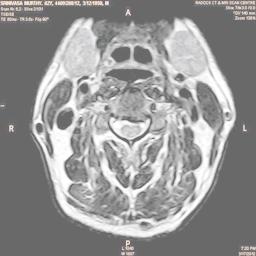}
}
\subfigure{
\includegraphics[height = 1.5 in, width = 1.4 in]{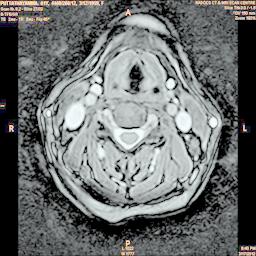}
}
\subfigure{
\includegraphics[height = 1.5 in, width = 1.4 in]{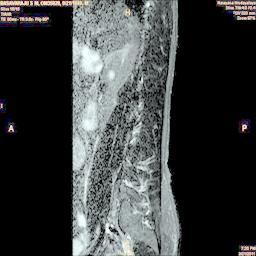}
}
\subfigure{
\includegraphics[height = 1.5 in, width = 1.4 in]{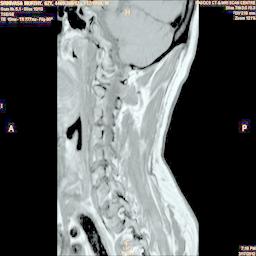}
}
\subfigure{
\includegraphics[height = 1.5 in, width = 1.4 in]{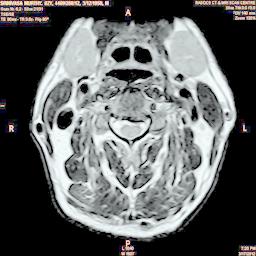}
}
\subfigure{
\includegraphics[height = 1.5 in, width = 1.4 in]{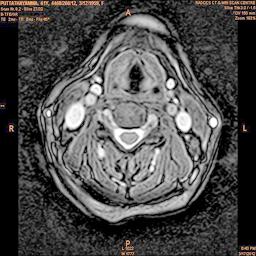}
}
\subfigure{
\includegraphics[height = 1.5 in, width = 1.4 in]{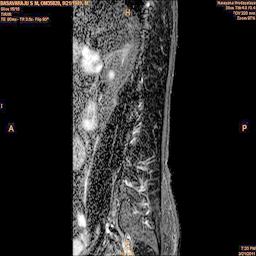}
}
\subfigure{
\includegraphics[height = 1.5 in, width = 1.4 in]{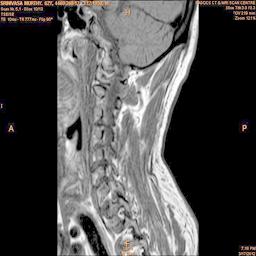}
}
\subfigure{
\includegraphics[height = 1.5 in, width = 1.4 in]{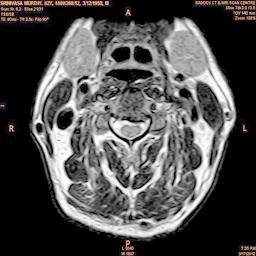}
}
\label{fig6}
\caption{Simulation Results of Spinal Cord Images. \textbf{First Row:} Original MRI of Thoracic Spine Showing Disc Herniation of Resolution $256\times256$ pixels. \textbf{Second Row:} Image Enhancement using Histogram Equalization Method. \textbf{Third Row:} Multiscale Retinex based Enhancement. \textbf{Fourth Row:} Proposed Multirate Multiscale Retinex based Enhancement}
\end{figure*}

In order to show the proposed method in more detail, the algorithm is tested with other test images. The first row of Fig. 6 shows the original MRI image of thoracic spine with different views. The second row of Fig. 6 shows the same image enhanced using histogram equalization. Third row of Fig. 6 presents the MSR based enhancement scheme. Fourth row of Fig. 6 indicates Chao's method of image enhancement. Finally, the reconstructed images using the proposed method is shown in fifth row of Fig. 6. Again, it can be seen from the results presented that the image enhancement using the proposed method is superior compared to other methods. This is evident from the image quality assessment presented in  Table 1 using AWE and DWE.   

\begin{table}
\centering
\small
\caption{Approximate and Detailed Wavelet Energy Metric for Quality Assessment of Results presented in Fig. 6, \textbf{Note:} The image quality assessment metric, AWE offers global image enhancement and DWE provides local enhancement details. In this work, priority is given for detail enhancement.}
\begin{tabular}{|l|c|c|c|c|c|c|c|c|}
\hline
&\textbf{AWE}& \textbf{DWE}& \textbf{AWE}& \textbf{DWE}& \textbf{AWE}& \textbf{DWE}& \textbf{AWE}& \textbf{DWE} \\
\hline 
Original Image& 98.20 &1.796  &97.02  &2.97 &98.80  &1.197 &98.91	&1.086    \\ 
\hline 
HE&99.98 &1.011  &99.88  &0.116 &99.84  &0.150 &99.75	&0.241    \\ 
\hline
MSR&98.62 &1.372  &98.70  &1.296 &99.52  &0.475 &99.00	&0.997    \\ 
\hline
Proposed&98.17 &1.825  &99.75  &2.252 &99.81  &0.607 &99.74	&1.176    \\ 
\hline
\end{tabular}
\label{table1}
\end{table}

\section{Conclusion}
The improved multiscale retinex algorithm based on multirate sampling scheme for spinal cord medical image enhancement application has been presented. The spinal cord image enhancement is achieved by applying MSR to each of five scaled versions of the Value component subsequent to contrast stretching operation. The speed of the proposed multiscale retinex algorithm has significantly improved, since the different sampled versions of Value components are processed in parallel. The reconstruction scheme adapted in this work is highly efficient compared with other methods. This is verified by conducting elaborate experiments on over a dozen varieties of spinal cord medical images. The reconstructed images are evaluated using an objective image quality assessment metric, wavelet energy in wavelet domain. The experimental results presented confirms that the proposed multirate multiscale retinex based image enhancement method offers high quality medical images. Research is in progress to extend the work for the enhancement of aerial images.




\begin{thebibliography}{1}

\bibitem{book01}
E.~R Davies, \emph{Computer and Machine Vision: Theory, Algorithms, Practicalities}, Academic Press, 2012.

\bibitem{book02}
Klaus D.~Toennies, \emph{Guide to Medical Image Analysis: Methods and Algorithms}, Springer, 2012.


\bibitem{journ01}
E. Vokurka, N. Thacker and A. Jackson, \emph{A Fast Model Independent Method for Automatic Correction of Intensity Non-uniformity in MRI Data}, Journal of Magnetic Resonance Imaging, Vol. 10, No. 4, pp. 550-562, 1999.

\bibitem{journ02}
P. Natarajan, N. Soniya, N. Krishnan, \emph{Fusion of MRI and CT Brain Images by Enhancement of Adaptive Histogram Equalization}, International Journal of Scientific and Engineering Research , Vol. 4, Issue 1, 2013.

\bibitem{journ03}
M. Sundaram, K. Ramar, N. Arumugam and G. Prabin, \emph{Histogram–Modified Local Contrast Enhancement for Mammogram Images}, International Journal of Biomedical Engineering and Technology, Vol. 9, No. 1, pp. 60-71, 2012.

\bibitem{journ04}
G. Cosnard, \emph{Tips and Traps in Spinal Cord Pathology}, Diagnostic and Interventional Imaging, 2012.

\bibitem{journ05}
Ronald Boet, Yu-Leung Chan, Ann King, Chung-Tong Mok and Wai-Sang Poon, \emph{Contrast Enhancement of the Spinal Cord in a Patient with Cervical Spondylotic Myelopathy}, Journal of Clinical Neuroscience, Vol. 11, No. 5, pp. 512-514, 2004.

\bibitem{journ06}
Daniel J. Jobson, Zia-ur Rahman and Glenn A. Woodell, \emph{A Multiscale Retinex for Bridging the Gap Between Color Images and the Human Observation of Scenes}, IEEE Transactions on Image Processing, Vol. 6, No. 7, pp. 965-976, 1997.

\bibitem{conf01}
Haifeng Wang and Yi Zhang, \emph{Image Enhancement Algorithm Using Brightness Preserving Multiple-Interval Histogram Equalization}, Proceedings of the $2^{nd}$ International Conference on Green Communications and Networks 2012 (GCN 2012), Vol. 2, pp. 647-654, 2013.

\bibitem{conf02}
Vladimir Todorov and Roumiana Kountcheva, \emph{Adaptive Approach for Enhancement the Visual Quality of Low-Contrast Medical Images}, Proceedings of the $2^{nd}$ Advances in Intelligent Analysis of Medical Data and Decision Support Systems, pp. 69-78, 2013.

\bibitem{conf03}
Qingyuan Meng, Deqian Bian, Mengfan Guo, Fengmei Lu and Dongpu Liu, \emph{Improved Multi-Scale Retinex Algorithm for Medical Image Enhancement}, Information Engineering and Applications, Vol. 154, pp. 930-937, 2012.

\bibitem{conf04}
Tamalika Chaira, \emph{Medical Image Enhancement Using Intuitionistic Fuzzy Set}, $1^{st}$ IEEE International Conference on Recent Advances in Information Technology (RAIT), pp. 54-57, 2012.

\bibitem{conf05}
K. P Indira and R. Rani Hemamalini, \emph{A Method for Contrast Correction and Enhancement For Medical Images using Wavelet Fusion}, In proceedings of International Conference on Computing and Control Engineering (ICCCE 2012), 2012.

\bibitem{conf06}
Chao An and Mei Yu, \emph{Fast Color Image Enhancement based on Fuzzy Multiple-Scale Retinex}, In 6th International Forum on Strategic Technology (IFOST 2011), Vol. 2, pp. 1065-1069, 2011.

\bibitem{conf07}
M. C Hanumantharaju, V. N Manjunath Aradhya, M. Ravishankar, and A. Mamatha, \emph{A Particle Swarm Optimization Method for Tuning the Parameters of Multiscale Retinex based Color Image Enhancement}, In Proceedings of the International Conference on Advances in Computing, Communications and Informatics, pp. 721-727, 2012.

\bibitem{conf1}
Dongni Zhang, Won-Jae Park, Seung-Jun Lee, Kang-A Choi and Sung-Jea Ko, \emph{Histogram Partition based Gamma correction for Image Contrast Enhancement}, 16th IEEE  International Symposium on Consumer Electronics (ISCE-2012), 2012.


\end{thebibliography}
%

\end{document}